\documentclass[sn-vancouver]{sn-jnl}

\jyear{2021}%

\theoremstyle{thmstyleone}%
\theoremstyle{thmstyletwo}%

\theoremstyle{thmstylethree}%
\usepackage{rotating} 

\begin{document}

\title[Resilient robot teams]{Resilient robot teams: a review integrating decentralised control, change-detection, and learning}


\author*[1]{\fnm{David M.} \sur{Bossens}}\email{d.m.bossens@soton.ac.uk}

\author[1]{\fnm{Sarvapali} \sur{Ramchurn}}\email{sdr1@soton.ac.uk}

\author*[1]{\fnm{Danesh} \sur{Tarapore}}\email{dst1m17@soton.ac.uk}

\affil[1]{\orgdiv{School of Electronics and Computer Science}, \orgname{University of Southampton}, \orgaddress{\city{Southampton}, \postcode{SO17 1BJ}, \country{UK}}}


\abstract{\textbf{Purpose of review:} This paper reviews opportunities and challenges for decentralised control, change-detection, and learning in the context of resilient robot teams.\\
\textbf{Recent findings:} Exogenous fault detection methods can provide a generic detection or a specific diagnosis with a recovery solution. Robot teams can perform active and distributed sensing for detecting changes in the environment, including identifying and tracking dynamic anomalies, as well as collaboratively mapping dynamic environments. Resilient methods for decentralised control have been developed in learning perception-action-communication loops, multi-agent reinforcement learning, embodied evolution, offline evolution with online adaptation, explicit task allocation, and stigmergy in swarm robotics. \\
\textbf{Summary:}
Remaining challenges for resilient robot teams are  integrating change-detection and trial-and-error learning methods, obtaining reliable performance evaluations under constrained evaluation time, improving the safety of resilient robot teams, theoretical results demonstrating rapid adaptation to given environmental perturbations, and designing realistic and compelling case studies.
}


\maketitle

\section{Introduction}
From monitoring wildlife \cite{Marques2012}, exploring dangerous terrains \cite{Roucek2020}, or collaboratively transporting items \cite{Montemayor2005}, robotic systems have the potential to transform society by performing tasks that are too dangerous, laborious, or repetitive for humans. In many such tasks, employing a \textit{robot team}, which consists of multiple robots, often yields significant advantages over employing just a single robot. For example, robot teams can monitor larger areas due to a larger spatial spread and provide robustness to failure due to redundancy (i.e. multiple robots performing the same or a similar subtask) \cite{Brambilla2013}.

In any given mission, robots in the team must control their own actuators and outgoing communications based on local sensory data and incoming communications. Decentralised control in this sense has been studied for multi-robot systems \cite{Farinelli2004,Yan2013}, which have fewer robots with more powerful hardware and high-complexity algorithms such as deep neural networks (e.g. \cite{Hu2021}), and swarm robotics \cite{Bayindir2016}, which is scalable in team size but comes with simplistic hardware and low-complexity algorithms such as ant-inspired algorithms (e.g. \cite{Dorigo2000}).

When robot teams are expected to become employed in long-term autonomy, they must be resilient to disruptions such as sensory-motor faults, weather conditions affecting the operating environment, or even adversarial cyber-attacks. The designer may anticipate some of these disruptions and hard-code their diagnoses and solutions; however, when there are complex or unanticipated disruptions, the robot team must detect these and adapt accordingly. While generic methods for change-detection \cite{Kotu2019,Yang2021} and transfer learning \cite{Lazaric2013b} have been applied in abstract machine learning problems, their application to robot teams comes with unique challenges such as communicating and integrating data from different robots' local error-prone observations and cooperative learning without incurring costs within the physical environment. 

With this context in mind, this paper reviews ongoing research on resilient robot teams within a decision process framework integrating decentralised control, change-detection, and learning. Section~\ref{sec: framework} first presents the integrated decision process framework and different prototype approaches within the framework. Section~\ref{sec: change-detection} reviews how robot teams can cooperate in detecting faults as well as changes in the surrounding environment. Section~\ref{sec: policy learning} reviews how decentralised control strategies can be learned for allowing robot teams to adapt to changed environments. Finally, Section~\ref{sec: conclusion} provides the main conclusions from the review along with future research directions.

\section{Integrating decentralised control, change-detection, and learning}
\label{sec: framework}
\begin{figure}[htbp!]
\centering
\includegraphics[width=1.0\textwidth]{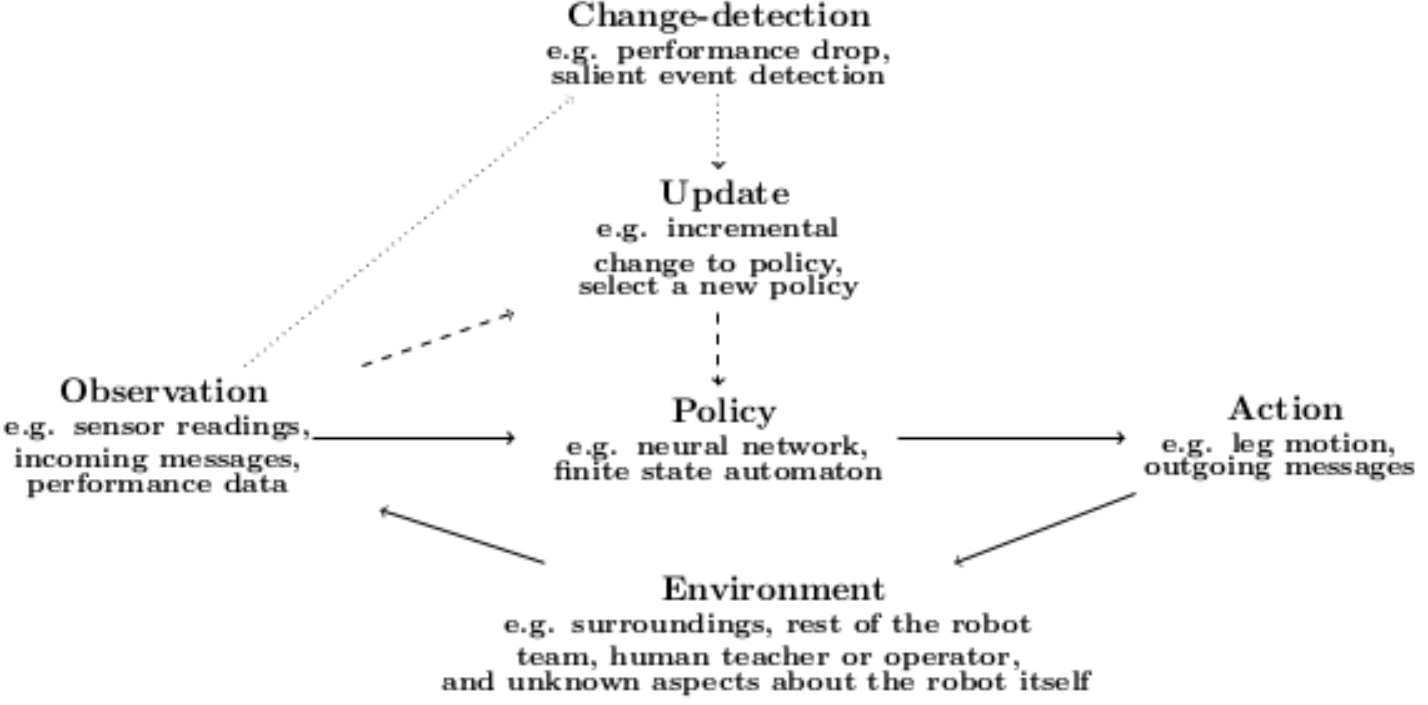}
\caption{Integrated decision process framework. The framework integrates decentralised control, change-detection, and learning into a single loop describing the decision problem faced by each robot in a robot team. Solid lines indicate the control cycle, dashed lines indicate the learning cycle performing policy updates periodically, and dotted lines indicate the optional change-detection cycle.} \label{fig: framework}
\end{figure}

From decentralised control to change-detection and adaptation to changes, robot teams face a challenging decision process. As illustrated in the integrated decision process framework shown in Figure~\ref{fig: framework}, each robot in the team must integrate its sensory readings and incoming communications to decide on which movements to take, which communications to send, and when and how to adapt to changes in the environment. To achieve this aim, each robot in the team uses a policy which selects an action to take at each control cycle, and performs change-detection and learning to adapt. Within our integrated decision process framework, the following section establishes a common vocabulary and key concepts for analysing the various methodologies for resilient robot teams.

\subsection{Communication}
Communication is an essential part of the decision-making process and comes in two forms, namely explicit communication and implicit communication \cite{Yan2013}. \textbf{Explicit communication} is an encoded message from a transmitter to receiver, and includes raw \cite{Li2018} or processed sensory readings \cite{Moratuwage2012}, tasks \cite{Dias2004}, and the performance of team members \citep{Parker2006}. Being communicated via wireless connections, they come with limitations such bandwidth, cost, and delays. In \textbf{implicit communication}, the sender manipulates the environment, for example by using gestures \cite{Bullard2020} or by placing objects in the environment as signs \cite{Dorigo2000}, and the receiver observes this directly via sensory readings. It avoids communication overload but may have more limited range, making it preferable for high robot densities.

When the robot-environment interaction changes, for instance due to sensory-motor faults or communication disturbances, the best communication strategy may change over time; in this case, learning how to communicate is an important capability. \textbf{Learning-how-to-communicate} approaches \cite{Peng2017,Bullard2020,Foerster2016a,Jiang2018} include aspects of communication into the policy to be updated during the learning cycle.

\subsection{Adaptation to changes}
To allow rapid adaptation to changes, the integrated decision process framework requires 1) a policy space that captures the desired behaviours in the policy space without needlessly increasing the search space; 2) a learning algorithm with strong empirical and theoretical support; and 3) a change-detection algorithm that balances genericity of application with specificity of diagnosis to inform the learning algorithm. 

Research in resilient robot teams has addressed these issues from different strategies, which we categorise based on four prototypes, namely diagnose-and-solve, trial-and-error, teacher-student, and robust-by-design, as summarised in Table~\ref{tab: prototypes}. In \textbf{diagnose-and-solve}, map a diagnose (e.g. localisation error) onto a specific ``repair'' (e.g. camera recalibration). This is the reasoning behind traditional fault-detection and fault-diagnosis methods \citep{Khalastchi2019} such as learning-based fault-diagnosis \citep{Parker2006}. 
In \textbf{trial-and-error}, the robots adapt to the environment by trying out different policies and evaluating their performance. Key approaches include multi-agent reinforcement learning \cite{Canese2021}, embodied evolution \cite{Ficici1999}, and offline evolution with online adaptation \citep[$\bullet \bullet$][]{bossens2021rapidly}. 
In \textbf{teacher-student}, the robots learn their policy from a teacher, typically using a large data set. The approach is typically based on imitation learning, i.e. by supervised learning from a data set of desired behaviours (e.g. \citep[$\bullet \bullet$][]{Hu2021}), where resilience can be provided by generalisation or by learning on a newly provided data set. The source of the data set varies but common examples are simulated trajectories generated by centralised expert controllers which have global state information \cite{Gama2020,Hu2021} and real-world video recordings of humans performing the desired behaviour \cite{Yu2018}.
In \textbf{robust-by-design}, each control cycle accounts for a range of environmental changes, for example by implicitly communicating occupancy map changes \cite{Dorigo2000} or by explicitly communicating the availability and capability of team members \citep{Parker1998}.

\begin{sidewaystable}
\centering
\caption{Adaptation prototypes within the integrated decision framework.} \label{tab: prototypes}
\begin{tabular}{l l p{4.0cm} p{4.0cm} p{4.0cm}}
\toprule
\textbf{Prototype} & \textbf{Policy space} & \textbf{Change-detection} & \textbf{Learning} & \textbf{Approaches}\\
\midrule
Diagnose-and-solve & domain-specific & diagnosis of a priori known faults & no learning or limited to generalising solutions to new faults & fault-detection and fault-diagnosis \\
Trial-and-error &  generic & generic change-detection & learning by trial-and-error  & multi-agent reinforcement learning, embodied evolution, and offline evolution with online adaptation  \\
Teacher-student & generic & new data provided by a teacher, often a human operator & imitation learning on provided data set  & learning perception-action-communication loops \\
Robust-by-design & domain-specific & implicit or explicit communication & frequent updates on desired task allocation & explicit task allocation, stigmergy in swarm robotics   \\
\bottomrule
\end{tabular} \\
\footnotesize{Columns indicate the prototypes and their various properties: \textbf{Policy space} indicates the space of possible policies for decentralised control; \textbf{Change-detection} indicates how change-detection is typically implemented; \textbf{Learning} indicates how the policy is adapted upon a detected environmental change. \textbf{Approaches} lists the different classes of algorithms within the adaptation prototype.}
\end{sidewaystable}

The above-mentioned approaches are analysed in more fine-grained detail in Section~\ref{sec: change-detection} and Section~\ref{sec: policy learning}. For now, we provide an overview of these approaches' applicability to different change types in Table~\ref{tab: change-types}. Changes in \textbf{team members} include changes in team size or in the capability or functionality of team members. Changes in \textbf{communication} include communication noise and the limited range of communication potentially cutting off some team members. Changes in the \textbf{transition dynamics or task requirements} include moving targets, obstructions, task complexity increases, and even mission objective changes. 

\begin{sidewaystable}
\centering
\caption{Change types and available methods.} \label{tab: change-types}
\begin{tabular}{p{4.0cm} | p{4.5cm} p{4.5cm} p{4.5cm}}
\toprule
\textbf{Change type} & \textbf{Examples} & \textbf{Change-detection} & \textbf{Policy learning} \\
\midrule
\textbf{Team members} &    team size change \newline sensory-motor or software fault \newline capability change   & Model-based fault-detection \cite{Parker2006,Christensen2007,Christensen2008}\newline
																   Feature-based fault-detection \cite{lau2011adaptive,Tarapore15a,tarapore2019fault}\newline
																   Synchronisation \cite{Christensen2009} \newline
																   Cryptography \cite{Ferrer2021}
																  &
																   Embodied evolution \cite{Ficici1999,Hart2018,Bredeche2012,Silva2017a,Winfield2011}  \newline
																   Offline evolution with online adaptation \citep{bossens2021rapidly,Bossens2020} \newline
																   Adaptive specialisation \citep{Parker1998,Emam2020,Emam2021} \newline
																   Explicit negotiation \cite{Dias2004,Gerkey2002} \newline
																   Ad hoc teamwork \cite{Stone2010}\newline 
																   Dynamic DCOPs \cite{Fioretto2018,Ramchurn2010} \\ \hline					   
\textbf{Communication} &  communication noise \newline local communication range & Model-based fault-detection \cite{Parker2006,Christensen2007,Christensen2008}  &  Perception-action-communication loops  \cite{Gama2020,Hu2021} \newline
			  										Independent Dec-POMDPs \citep{Omidshafiei2017} \newline
			  										Networking Dec-POMDPs \citep{Sukhbaatar2016,Jung2021}   \\ \hline
\textbf{Transition dynamics \newline or task requirements} & dynamic targets \newline obstructions \newline task complexity increase \newline mission objective change  & Task identification \cite{Fifty2021,Lomonaco2020,Milan2016} \newline Dynamic and multi-vehicle mapping \cite{Saeedi2014,Tipaldi2013,Moratuwage2012} \newline multi-anomaly detection and tracking \cite{Saldana2015,Li2018,Salam} & Multi-task multi-agent reinforcement learning \cite{Omidshafiei2017}  \newline Embodied evolution \cite{Ficici1999,Bredeche2012,Silva2012,Hart2018,Silva2017a,Winfield2011}    \newline  Offline evolution with online adaptation \citep{bossens2021rapidly,Bossens2020} \newline Dynamic DCOPs \cite{Fioretto2018,Ramchurn2010}\\
\bottomrule
\end{tabular}
\end{sidewaystable}

\section{Change-detection}
\label{sec: change-detection}
Although the topic of change-detection is broad, we focus on the unique opportunities and challenges in the context of resilient robot teams. In particular, we focus on identifying faults in team members and detecting and tracking changes in the surroundings. The former follows the diagnose-and-solve prototype while the latter often requires more generic adaptation methods.

\subsection{Detecting faults in team members}
At any time in their mission, members of a robot team may experience faults. Robots within a team may detect faults in themselves, or endogenous fault-detection. However, exogenous fault-detection, in which robots detect faults in each other based on data collected from the different team members, makes full use of the team's joint capability to provide comparably higher efficiency and robustness. We distinguish here between four methods to detecting faults in team members, with an emphasis on exogenous fault-detection. The methods vary in their assumptions, genericity, and scalability, namely model-based fault-detection, feature-based anomaly detection, synchronisation, and cryptographic authentication.

Model-based methods model faults at design-time and detect them at run-time. Learning-based fault diagnosis \cite{Parker2006} pre-determines the causal links between symptom, fault, and solution for a priori known faults. It uses case-based reasoning to identify new cases by comparing them to the existing database of causal links (containing pre-defined or previously encountered faults) based on the most probable symptoms that are relevant for task completion. If the new case cannot be classified in terms of pre-existing causal links, then a human operator must communicate the recovery solution. Another approach analyses the difference of the observed dynamics to the dynamics expected under normal or faulty conditions. This method was first pursued in endogenous fault-detection using Kalman filters as models for linear dynamical systems  \citep{Roumeliotis1998,VanEykeren2011} or neural network models for more generic applicability \citep{Skoundrianos2004a,Terra2001}. Later, in a method suitable for endogenous and exogenous fault-detection, Christensen et al. \cite{Christensen2007,Christensen2008} use time-delay neural networks to model faulty sensors or actuators by fitting sensory-motor data coming from trajectories under the fault and compare these to observed sensory-motor data observed from other robots. 
The expressivity of neural networks makes the approach ideal for detecting complex symptoms of failure. Unlike learning-based fault-diagnosis, the approach is sensitive to an arbitrary threshold parameter for the neural network classification and does not lend well to high-level behaviours such as path planning, localisation, and following behaviours. Scalability is a question mark, as the methodology was shown on the leader-follower task which only contains two robots.

A generic and scalable approach to exogenous fault detection is to perform distance- or probability-based anomaly detection based on team members' feature-vectors which record features of interest received by explicit communications from team members (e.g. sensory-motor trajectories). In Lau et al. \citep{lau2011adaptive}, robots check from communicated data whether or not other robots experience the same change and if not a fault is present. Their results find that the receptor density algorithm \citep{Owens2013}, a non-parametric kernel-based technique inspired by T-cell receptors, significantly outperformed 4 other statistical tests. In a related immune system inspired method, faults are detected by individual robots applying a cross-regulation model \cite{Carneiro2007,Leon2001} on behavioural feature vectors of neighbouring robots and then voting on which robots are faulty  \cite{Tarapore15a,tarapore2019fault}; the method has been evaluated on a variety of tasks and environmental changes (including physical experiments \citep[$\bullet$][]{tarapore2019fault}), showing tolerance to changing swarm behaviours as well as individual robots behaving anomalously. A potential downside of the immune-inspired approach is that rather than faults, differences in behavioural feature vectors might reflect localised environmental conditions such as reduced friction or the presence of obstacles.

Synchronisation \cite{Christensen2009} assumes robots are required to send out physical signals (e.g. LED light flashes) and require them back from their neighbouring team members; if a neighbouring robot does not synchronise, it must have a fault of some sort (e.g. light sensors, light actuators, or motion actuators). Unfortunately, the approach is rather hardware-specific and does not easily extend to fault diagnosis.

Cryptographic approaches focus specifically on identifying team members that have been compromised by a cyber-attack. The approach by Ferrer et al. \cite[$\bullet $][]{Ferrer2021} relies on a Merkle tree data structure, which ensures that each robot must share cryptographic proofs to verify their integrity before cooperating on the mission. While the approach is only applicable to security risks, it can detect such changes even when the robot under attack does not demonstrate any observable behavioural differences.

\subsection{Detecting changes in the surroundings}
Compared to traditional change-detection in machine learning, robot teams are distributed in space and can actively reposition themselves to make sense of the surrounding environment, which has special relevance for applications such as disaster recovery, search-and-rescue, and environmental monitoring. 

One domain of interest is detecting and tracking anomalies in the environment. In the approach by Saldana et al. \cite{Saldana2015}, team members can sense anomalous regions directly in the environment, explicitly communicate their observations to each other, and try to surround multiple existing anomalies while exploring the map to find new anomalies. The approach by Li et al. \cite{Li2018} formulates a dynamic optimisation problem in which the optimum to be tracked is the maximum (or minimum) on a particular feature of interest. Robots function as distributed searchers that occupy promising areas more densely and that communicate each others' measurements. Unfortunately, the approach was only demonstrated on toy function optimisation problems, so a physical robotics demonstration could make the approach more convincing. For physical fields such as oceans, Salam et al. \cite{Salam} present an algorithm for estimating the full state of a dynamic process based on robots within a team communicating their local observations of the quantity to be tracked (e.g. concentration of particles, temperature) and then recomputing the updated system dynamics. The approach was demonstrated to have high accuracy compared to radial basis function interpolation in a physical water tank.

Similarly, due to their distributed active sensing capabilities, cooperative techniques are also being investigated for simultaneous localisation and mapping (SLAM) within robot teams \cite{Saeedi2014}. Traditional multi-vehicle SLAM has two main disadvantages, namely that forming joint map from the local observations has high computation and communication overhead and that the objects in the map formed by SLAM are assumed to be static. To counter these issues, one can track static and dynamic features in the map using dynamic occupancy grids \cite{Tipaldi2013} and communicate local maps rather than raw sensory data  \cite{Moratuwage2012}.

\section{Policy learning for decentralised control}
\label{sec: policy learning}
This section describes how policies can be learned for allowing robot teams to adapt to changed environments. We distinguish broadly between six approaches: one teacher-student approach called learning perception-action-communication loops; three trial-and-error approaches, namely multi-agent reinforcement learning,  embodied evolution, and evolution with online adaptation; and two robust-by-design approaches called explicit task allocation and stigmergy in swarm robotics.
\subsection{Learning perception-action-communication loops}
Incorporating perception-action-communication loops into graph neural networks (GNNs) is a recent solution to decentralised control and learning-how-to-communicate \cite{Gama2020,Hu2021}. In such works, the GNN takes as input a communication graph, which represents robots as nodes, communication links as edges, and communication delays as their distances. The GNN is learned via imitation learning, a teacher-student approach in which the policies of the robots are updated by supervised learning on trajectories collected by the teacher (e.g. a human operator who has evaluated the best joint policy for the new situation). As in Hu et al. \citep[$\bullet \bullet$][]{Hu2021}, which assumes each robot has a transceiver, the graph neural network can learn to process incoming messages and send outgoing messages, and integrate this with local visual observations, resulting in a system that is robust to visual degradation as well as changes in team size and communication graph topology. A key challenge for imitation learning is that expert knowledge of the teacher will not always be available. Although domain adaptive imitation learning \citep{Kim2020} has explicitly targeted imitation learning across different tasks, this still relies on the availability of expert data sets for each task. Alternatively, few-shot imitation methods \citep{Yu2018} provide an option for imitation learning with limited data but so far has been studied only in single-robot contexts.
\subsection{Multi-agent reinforcement learning}
Multi-agent reinforcement learning (MARL) is a relatively mature field which is grounded in the sound theoretical foundations of Markov decision processes with various analytical convergence results (e.g. \citep{Zhang2018}). In MARL, different agents receive observations of the environment, perform an action, and receive rewards indicating how desirable the behaviour is, and by trial-and-error the agents learn to collectively optimise the long-term cumulative reward. In MARL, resilience is conceptualised as learning a near-optimal policy in response to a change in the decision process. 

Among the variety of MARL frameworks, decentralised partially observable Markov decision processes (Dec-POMDPs; \cite{Oliehoek2013}) are of particular relevance to robot teams, as they integrate partial observability (e.g. due to limited sensory observations) into decentralised control. While there have been a variety of powerful alternative frameworks, such as  Communicative Multi-agent Team Decision Problems \cite{Pynadath2002}, these have not been as widely investigated. Recent works within Dec-POMDPs has examined multi-task learning and robustness to communication failures. Decentralised hysteretic deep recurrent Q-network (Dec-HDRQN) \citep{Omidshafiei2017} uses hysteretic Q-learning \citep{Matignon2007}, in which agents update their own independent Q-table with lower learning rate (without communication) when this is likely due to other agents' suboptimal actions, while using deep recurrent Q-network \citep{Hausknecht2015} for partial observability and policy distillation \citep{Rusu2016} for improved generalisation. Dec-HDRQN allows learning many tasks within a single policy without requiring explicit task identification, as demonstrated on tasks with different grid sizes and transition dynamics in a grid-world domain. While independent decentralised control has benefits when communication channels are unreliable or faulty, the strategy is suboptimal. Instead, ``networking'' Dec-POMDPs can additionally incorporate communication amongst the different agents; popular approaches include sharing parameters and then forming a consensus \citep{Zhang2018}, factorisation \citep{Sunehag2018,Rashid2018}, and communication protocols learned concurrently with the policy \citep{Sukhbaatar2016,Foerster2016}. Among these, CommNet \citep{Sukhbaatar2016} accounts for dynamic variation of the type and number of agents communicating and has been demonstrated for energy sharing in multi-UAV systems for distributed data processing \cite[$\bullet \bullet$][]{Jung2021}. 

Assuming a change is detected in the decision process (e.g. using task identification methods \cite{Fifty2021,Lomonaco2020,Milan2016}), MARL systems require extensive experience to learn a suitable policy if it is not already available from prior training. When a high-fidelity simulator is available, this may not be problematic and in this case one may also consider improving performance by applying centralised training with decentralised execution in the online phase (e.g. \cite{Foerster2018,Gupta2017}). Alternatively, for rapid adaptation, (theoretical or empirical) demonstrations of sample complexity are required rather than the asymptotic convergence guarantees for model-free MARL (e.g. methods based on Q-learning \cite{Watkins1992}).

\subsection{Evolution}
Evolutionary algorithms (EAs) mimic natural evolution to generate a diverse population of genomes and progressively select them for fitness over subsequent generations. While it is popular to evolve policies with EAs -- for example, using NEAT \cite{Stanley2002} to evolve the weights and topology of neural networks -- in the context of resilient robot teams we mainly find two dominant approaches, namely embodied evolution and offline evolution with online adaptation.
\subsubsection{Embodied evolution}
Embodied evolution \cite{Ficici1999} evolves robots within their physical environment, which helps to avoid the ``simulation-reality gap'' \citep{Jakobi1995} and to adapt to changes in the real world. Each robot executes its policy based on its current genotype but the policy has a limited lifespan based on a ``virtual energy'' quantity which represents the robot's own performance estimate and which (in some implementations) improves the chances of reproduction. For reproduction, robots communicate with each other by broadcasting their own genotype, some genes of which are then integrated into the receiving robots' genotypes. 

Unfortunately, this process of adaptation can take hours \citep{Bredeche2012,Silva2012}. Functional diversity can also be developed by using a quality-diversity approach to embodied evolution \citep{Hart2018}, which can be used as an implicit task allocation. With focus on real robot swarms, Silva et al. (2017) \citep{Silva2017a} demonstrate odNEAT \cite{Silva2012}, an online variant of NEAT, as being resilient to simulation-to-reality transfer, task change, and fault injection.

Rather than evolving genotypes, embodied evolution may also operate on memes, which are cultural utterances defined in robot teams as ``contiguous sequences or packages of movements, or sounds, copied from one robot to another, by imitation'' \citep{Winfield2011}. This approach uses implicit communication by observing each others' behaviours with sensory observations, which reduces communication overhead and more directly searches the policy space but is limited by observability.

\subsubsection{Offline evolution with online adaptation}
\label{sec: offline-online}
Embodied evolution is time-consuming, so another approach is to first evolve a large archive of policies offline and then perform rapid online adaptation by searching this policy space after a change (e.g. performance drop) is detected. In this context, offline evolution is based on quality-diversity  (QD) algorithms (e.g. \citep{Mouret2015b,LehmanStanley2011}), which evolve an archive of behaviourally diverse and high-performing solutions, while online adaptation is based on Bayesian optimisation and has been studied primarily in single-robot adaptation to damaged actuators \citep{Cully2015b,Dalin2019a,Papaspyros2016}. 

Two recent methods have expanded the approach to be applied in robot teams and in a wider set of environmental changes. Swarm Map-based Optimisation Decentralised (SMBO-Dec) \citep[$\bullet \bullet$][]{bossens2021rapidly} forms a Gaussian process model for each subgroup within the team based on local environmental conditions. Different robots in the team function as different workers in an asynchronous batch-based Bayesian optimisation,  which helps to speed up the search while avoiding to sample similar behaviours simultaneously. The empirical success was demonstrated by 80\% performance improvements within a mere 30 evaluations on a large variety of perturbations, including food scarcity and different sensory-motor faults. Due to subgrouping the team based on local conditions, the approach combines naturally with diagnoses of change-detection algorithms, although this option has not been explored yet. Quality-Environment-Diversity (QED) \citep{Bossens2020} evolves behavioural diversity based on the type of environments that the policies solve. Since QED archives represent solutions to different environments, they may be more efficient for online adaptation than traditional QD archives, especially when robots can provide information on their current environment.

\subsection{Explicit task allocation}
Explicit task allocation explicitly defines sub-tasks within a mission, and assigns different robots to them based on task priority and robot capability. This reduces the complexity of the mission but limits behavioural flexibility and requires design-time knowledge as well as frequent explicit communication during run-time. The approach is robust-by-design as task allocation dynamically accounts for the unique and changing task priorities and robot capabilities.

In adaptive specialisation, robots change their roles if they detect insufficient task progress.
Representative of this approach are ALLIANCE \citep{Parker1998}, in which each robot activates a high-level behaviour from its set based on how incoming communications affecting its internal motivations, and data-driven adaptive multi-robot task allocation \cite[$\bullet \bullet$][]{Emam2021}, which makes low (resp. high) performance of robot $i$ on 
task $j$ being followed by a reduction (resp. increase) in the task specialisation $s_{ij}$. The latter approach demonstrated successful task re-allocation in a team composed of ground robots and quadcopters in the  Robotarium (see \cite{Pickem2017}) after parts of the space became no-go or no-fly zones  \cite{Emam2020,Emam2021}. Downsides of adaptive task specialisation are the sensitivity to hyperparameters (e.g. time until impatience), the frequent modelling, communication, and global information required to evaluate task progress, and the model assumptions (e.g. control-affine systems). 

Ad hoc teamwork \cite{Stone2010} considers the related problem of allocating a subset of agents from a pool to participate in solving a particular task, based on their capabilities in the domain from which the task is sampled. While traditionally these capabilities are pre-defined, recent work integrates convolutional neural network based change point detection of capability changes in non-stationary agents \cite{Ravula2019}.

In explicit negotiation, agents bid for their preferred tasks, typically based on the contract-net protocol \citep{Smith1980}. Within this approach, TraderBots \cite{Dias2004} demonstrated adaptive task allocation on communication failure, partial and complete robot failure, and the reintroduction of a once failed robot, and MURDOCH \cite{Gerkey2002} demonstrated adaptation to new tasks as well as individual robot failures.

Task allocation can be cast as a distributed constrained optimisation problem (DCOP) \cite{Fioretto2018}, by allocating each agent in the team to one or more mission variables to optimise the mission's objective, which is composed of different cost functions on subsets of mission variables. Of particular interest is the dynamic DCOP, which allows the DCOP to change over time. Within this framework, it is for example possible to solve search-and-rescue missions with robustness to run-time addition or removal of tasks, defined as victim locations that are reachable within deadline \cite{Ramchurn2010}. Dynamic DCOPs do come with an increase in explicit communication and computation.

\subsection{Stigmergy in swarm robotics}
Stigmergy is a form of implicit communication in which one agent drops signs in the environment to pass information to other agents, an approach which is scalable to the large team sizes found in swarm robotics.  Most traditionally, ant colony algorithms leave pheromone trails in the environment to signal where other agents should come \cite{Dorigo2000}. The approach is mostly applicable to foraging or search tasks, where it can account for road blockades, making the approach robust-by-design. However, physically implementing pheromones is always a challenge in real-world applications \cite{Zedadra2017} as the physical signs, when dropped \textit{en masse} should not incur environmental costs, interference with human activities, etc. Recent work has explored the use of light sources but requires either an overhead camera and light projector \cite{Hunt2019} or locally placed photochromic materials \cite{Salman2020}.

\section{Conclusions}
\label{sec: conclusion}
Integrating decentralised control, change-detection, and learning is an important challenge facing robot teams for resilience in real-world applications such as search-and-rescue, environmental monitoring, exploration, and pickup-and-delivery. 

Change-detection in robot teams comes with unique opportunities and challenges. Team members can cooperatively identify faults in each other as well as actively seek, identify, and track anomalies in the surrounding environment. They share a limited space which comes with localisation and communication constraints (e.g. deadlocks and communication overload), and they must seek to diagnose the faults and provide a solution in a generic manner. 

Robot teams face a challenging decision process with decentralised control of actuators and communication channels. Perception-action-communication loops can be learned such that incoming and outgoing messages are processed by the policy of the robot. These rely on imitation learning, which requires an expert to provide new data when the environment changes. MARL is pursued for robot teams in Dec-POMDPs, which has been recently investigated in the context of multi-task RL to improve generalisation to new environments. Despite only demonstrating asymptotic convergence (rather than transfer and sample efficiency), the existence of theoretical guarantees distinguishes the MARL framework from other frameworks. Embodied evolution evolves the robots directly in the physical environment which avoids the simulation-reality gap but can take many hours to achieve adaptation. The recently emerging field of offline evolution with online adaptation achieves rapid adaptation across a wide range of perturbations by using collaborative learning to update performance priors from behaviours evolved offline. Explicit task allocation breaks the mission down into smaller subtasks and allocates team mambers to them, which simplifies the adaptation problem but reduces the flexibility of the team by the pre-defined roles. Finally, stigmergy is based on dropping signs in the environment, which scales favourably with team size but is challenging to implement in physical robot teams.

While the above has addressed resilience to a wide array of change-types, including team members' capabilities or functionalities, limitations of communication, and environmental dynamics, we still foresee significant challenges in the field of resilient robot teams. First, the integration of change-detection and trial-and-error methods has not been widely explored; to start the adaptation process within a subset of environments, inspiration could be drawn from task identification methods \cite{Fifty2021,Lomonaco2020,Milan2016}. Second, there is a need for methods to obtain reliable performance evaluations without requiring too much evaluation time; there, the evaluation time could be large for promising solutions and small for clearly unsuccessful solutions, similar to the use of virtual energy in embodied evolution. Third, safety is still widely overlooked in the context of robot teams. A promising avenue is to extend single-agent methods for safe reinforcement learning (e.g. shielding \citep{Alshiekh2018}) and offline evolution with safe online adaptation  (e.g. map-based constrained Bayesian optimisation \citep{Papaspyros2016}) to the decentralised multi-agent setting. Indeed, shielding, which replaces any unsafe actions for a given state with a safe action, has recently been extended to factorised shielding \citep{ElSayed-Aly2021}, which maintains shields for subsets of agents based on a factorisation of the state space. Fourth, there is the challenge of designing algorithms for adaptive robot teams with formal proofs on their sample complexity under various classes of environmental perturbations. Fifth, since realistic applications are ultimately the driver of research in robotics, the design and implementation of convincing case studies is of critical importance. Such studies will highlight new challenges and opportunities, as well as demonstrate the trustworthiness of resilient robot teams to key players in industry and government.

\section*{Acknowledgements}
This work has been supported by the Engineering and Physical Sciences Research Council under the New Investigator Award grant, EP/R030073/1, and the UKRI Trustworthy Autonomous Systems Hub, EP/V00784X/1.

\section*{Compliance with Ethics Guidelines}
\subsection*{Conflict of Interest}
The authors declare that they have no conflict of interest.
\subsection*{Human and Animal Rights and Informed Consent}
This article does not contain any studies with human or animal subjects performed by any of the authors.


\bibliography{sn-bibliography}


\end{document}